\documentclass[sigconf]{acmart}
\AtBeginDocument{%
  \providecommand\BibTeX{{%
    \normalfont B\kern-0.5em{\scshape i\kern-0.25em b}\kern-0.8em\TeX}}}
    
\setcopyright{acmcopyright}

\copyrightyear{2023}
\acmYear{2023}
\setcopyright{rightsretained}
\acmConference[WWW '23]{Proceedings of the ACM Web Conference 2023}{April
30-May 4, 2023}{Austin, TX, USA}
\acmBooktitle{Proceedings of the ACM Web Conference 2023 (WWW '23), April
30-May 4, 2023, Austin, TX, USA}
\acmDOI{10.1145/3543507.3583862}
\acmISBN{978-1-4503-9416-1/23/04}

\usepackage[shortlabels]{enumitem}
\usepackage{caption}
\usepackage{subcaption}
\usepackage{graphicx}
\usepackage{cleveref}
\usepackage{adjustbox}
\usepackage{booktabs}
\usepackage{multirow}
\usepackage{lipsum}
\usepackage{balance}

\usepackage{color,soul}

\graphicspath{{figures/}}

\usepackage{geometry}
\newcommand{\para}[1]{\smallskip\noindent\textbf{#1}}
\newcommand{\best}[1]{\textit{\color{blue}{\textit{\textbf{#1}}}}}
\newcommand{\vn}[1]{\mathsf{#1}}

\begin{document}

\title{Interpreting wealth distribution via poverty map inference using multimodal data}
\titlenote{\color{red}Please cite the WWW'23 version of this paper}

\author{Lisette Espín-Noboa}
\email{EspinL@ceu.edu}
\affiliation{%
  \institution{Central European University}
  \country{}
}
\affiliation{%
  \institution{Complexity Science Hub Vienna}
  \country{}
}

\author{János Kertész}
\email{KerteszJ@ceu.edu}
\affiliation{%
  \institution{Central European University}
  \country{}
}
\affiliation{%
  \institution{Complexity Science Hub Vienna}
  \country{}
}

\author{Márton Karsai}
\email{KarsaiM@ceu.edu}
\affiliation{%
  \institution{Central European University}
  \country{}
}
\affiliation{%
  \institution{R\`{e}nyi Institute of Mathematics}
  \country{}
}

\renewcommand{\shortauthors}{Espín-Noboa, et al.}
\begin{abstract}
Poverty maps are essential tools for 
governments and NGOs to track socioeconomic changes and adequately allocate infrastructure and services in places in need. 
Sensor and online crowd-sourced data combined with machine learning methods have provided a recent breakthrough in poverty map inference.  However, these methods do not capture local wealth fluctuations, and are not optimized to produce accountable results that guarantee accurate predictions to all sub-populations. %
Here, we propose a pipeline of machine learning models to infer \textit{the mean and standard deviation of wealth} across multiple geographically clustered populated places, and illustrate their performance in 
Sierra Leone and Uganda.
These models leverage seven independent and freely available feature sources based on satellite \textit{images}, 
and \textit{metadata} collected via online crowd-sourcing and social media. %
Our models show that %
combined metadata features are the best predictors of wealth in rural areas, outperforming image-based models, %
which are the best for predicting the highest wealth quintiles. 
Our results %
recover the local mean and variation of wealth, and correctly capture the positive yet non-monotonous correlation between them. %
We further demonstrate the capabilities and limitations of model transfer across countries and the effects of data recency and other biases.
Our methodology provides open tools to build towards more transparent and interpretable %
models to help 
governments and NGOs to
make informed decisions 
based on data availability, urbanization level, and poverty thresholds. %
\end{abstract}

\begin{CCSXML}
<ccs2012>
  <concept>
      <concept_id>10010147.10010257.10010293.10010294</concept_id>
      <concept_desc>Computing methodologies~Neural networks</concept_desc>
      <concept_significance>300</concept_significance>
      </concept>
  <concept>
       <concept_id>10010405.10010481.10010487</concept_id>
       <concept_desc>Applied computing~Forecasting</concept_desc>
       <concept_significance>500</concept_significance>
       </concept>
   <concept>
       <concept_id>10002951.10003227.10003236.10003237</concept_id>
       <concept_desc>Information systems~Geographic information systems</concept_desc>
       <concept_significance>300</concept_significance>
       </concept>
    <concept>
       <concept_id>10002951.10003227.10003236.10003101</concept_id>
       <concept_desc>Information systems~Location based services</concept_desc>
       <concept_significance>100</concept_significance>
       </concept>

 </ccs2012>
\end{CCSXML}

\ccsdesc[300]{Computing methodologies~Neural networks}
\ccsdesc[500]{Applied computing~Forecasting}
\ccsdesc[300]{Information systems~Geographic information systems}
\ccsdesc[100]{Information systems~Location based services}

\keywords{high-resolution spatial inference, poverty maps, machine learning, deep learning, satellite images, online crowd-sourced data}

\maketitle
\section{Introduction}

The first Sustainable Development Goal (SDG) set by the United Nations is to eradicate poverty by 2030~\cite{UNpoverty}. Although fewer people were living in extreme poverty around the world by 2018, the decline in poverty rates has slowed down ever since. This stagnation was partly due to the COVID-19 pandemic, but the ongoing impact of political turmoils, wars, and climate catastrophes set further barriers for progress in this direction
~\cite{WBpoverty}. 
Traditional data collection techniques fail to follow the effects of such rapidly changing circumstances, therefore, new data collection and analysis techniques
are required.
Further, the identification of places in need requires rapid, flexible and precise inference to inform the adequate allocation of resources, which are often misplaced due to coarse-grained and outdated statistics provided by census and survey data.

The fast %
penetration of mobile phones~\cite{adam2018icts,areppim2017} 
has enabled the collection and use of big data for social good. For instance, mobile phone call detailed records (CDR) and mobile airtime payment transactions have been used to infer several socioeconomic indicators%
~\cite{dong2014inferring,gao2019computational,cruz2021estimating}, which in turn have been applied to map socioeconomic effects on the structure and dynamics of the underlying social networks~\cite{eagle2010network,leo2016socioeconomic,leo2018correlations}. One main limitation of CDRs, however, is that mobile phone data is proprietary and access is often granted via purchase or partnerships. 
On the contrary, the emergence of Web 2.0 technologies has opened new avenues for collecting and sharing online annotations and digital traces~\cite{hudson2009mapping, rafaeli2019digital}. Further, open data initiatives have strengthen collaborations between industry, academia, and the public sector~\cite{fbgood}, which have made it easier to study social, economic, and environmental issues through the estimation of socioeconomic indicators from online data~\cite{llorente2015social,aletras2018predicting, abitbol2018location, abitbol2021socioeconomic}.

In recent years, satellite imagery has attracted great attention as a mean of inferring high-resolution poverty maps. As a first attempt, nightlight intensity of places has been shown to be a good estimator of economic activity~\cite{ghosh2010shedding}, especially for non-extreme poor areas~\cite{pinkovskiy2016lights}. 
Moreover, they can even capture household consumption of the extreme poor when combined with daylight satellite images~\cite{jean2016combining}.
Other approaches leverage the visual qualities of street-level imagery and daylight satellite images to detect objects (e.g., vehicles, infrastructure, and terrain) and use them as proxies of wealth of neighborhoods~\cite{abitbol2020interpretable,ayush2021efficient,uzkent2021gen}.
In principle, any geospatial dataset can be used to build poverty maps from their spatial correlations with socioeconomic indicators. For instance, advanced deep learning methods trained on multiple features (e.g., satellite images together with population density maps) %
provide scalable and improved predictions of wealth at high-resolution for low- and middle-income countries around the globe~\cite{lee2020high,Chie2022micro}.

While these methods represent a great advance towards scalable, time-variant and fine-grained poverty maps, they commonly optimize performance over representativeness, and depend strongly on the availability of all data sources. As a result, they might perform poorly in countries with scarce data and fall short in accuracy guarantees necessary for policy makers. In addition, most of these methods use machine learning algorithms with good performance conditional to specific data engineering and parametrization choices. This limits their generalization and transferability potential, especially in countries in emergency undergoing rapid demographic and environmental changes. Although the parametrization of these models is well documented, their robustness against data scarcity, skewness, and time is usually not evaluated or not transparent. 

Here, we bridge this gap by performing a systematic study to better understand the capabilities of existing methods and feature sources, which to date have mostly been studied in isolation. 
Further, we propose regression models that combine features from seven freely available sources to show their overall performance and their strength at the intersection between socioeconomic classes and types of settlements.
We train our models to infer not only the mean wealth of places, but also their standard deviation to give a more precise view of how wealth is distributed across households within each populated place. 
In line with these objectives, we frame our analysis around the following research questions.

\begin{enumerate}[start=1,label={\upshape\bfseries RQ\arabic*:},wide = 15pt, leftmargin = 4.4em]
    \item What type of model and features are best at predicting both the mean and standard deviation of wealth? 
    \item Is the goodness-of-fit of models consistent across types of settlements and socioeconomic classes?
    \item How broadly is wealth distributed in each place? Is the best model able to capture the correlation between the mean and standard deviation of wealth? 
    \item What is the trade-off between data availability and geographical model transfer? %
\end{enumerate}

We showcase performance in Sierra Leone (SL) and Uganda (UG), two Sub-Saharan African countries characterized by extreme poverty~\cite{WBpoverty,UNpoverty}.
Using the last two household surveys conducted by the Demographics and Health Program (DHS)~\cite{dhs}, we compute the international wealth index (IWI)~\cite{smits2015international} of localized population clusters in these countries. In addition, we extract 173~metadata- and 784~image-based features for each location, at no cost, from four data providers: %
Google, Meta (Facebook), OpenStreetMap, and OpenCelliD. 
The image features refer to embeddings extracted from daylight satellite images while the metadata features include demographics, mobility, population density, nightlight intensity, mobile communication antennas, and ground infrastructure.
We found that inference performance varies widely across models, feature sources, and geography suggesting that wealth and poverty have different characteristics across different countries. For instance, population density is the best predictor in SL while nightlight intensity is the best in UG. In both cases, however, mobility features are the second best predictors. Moreover, the combination of all metadata-based features outperforms the image-based and individual-source features in both countries. Interestingly, geographical model transferability pays-off when features are partially missing. 
Finally, while no model predicts wealth equally well at the intersection between socioeconomic classes (e.g., poor, rich) and types of settlements (i.e., urban, rural), we found that, there is a potential in combining multiple models to increase performance.

Our analysis sheds light on how to use sensor and online crowd-sourced data to reliably infer high-resolution poverty maps.
To ensure transparency and reproducibility, we make our code and results openly available~\cite{espin2022codefinal},
and share an online visualization tool~\cite{yang2022vizfinal} to interactively demonstrate that our inferred high-resolution poverty maps can be used to identify places in need.

\section{Related work}

Closest to our work, 
\cite{Chie2022micro, lee2020high} %
combine multiple feature sources to predict high-resolution poverty maps using wealth as a proxy of poverty. In terms of model architecture, we build upon~\cite{lee2020high, jean2016combining} by adding more features into the pipeline and combining two regressor models, one based on images and the other based on metadata (tabular data). Note that the advantage of having multiple features is that the limitations of some sources can be overcome by the strengths of others. For example, models can learn correlations between features characterizing the same location but coming from different sources~\cite{rahate2022multimodal}.
In terms of analysis, only ~\cite{Chie2022micro} goes beyond the overall performance of the model and assess feature importance, and measures the goodness-of-fit of geographical model transferability and different sampling methods. 
Compared to this work, we additionally show under which circumstances geographical model transferability pays-off given the available features. Further, we compare the performance of multiple models and show their weaknesses and strengths at the intersection between socioeconomic classes and types of settlements. Importantly, unlike in any other work, we infer not only the mean but the standard deviation of the distribution of wealth in populated places; in this way we provide information about the level of socioeconomic diversity at each location. For a profound review on the potential of combining new kinds of data with artificial intelligence (or machine learning algorithms) to achieve one or multiple SDGs, see~\cite{gao2019computational, hidalgo2020digital, palomares2021panoramic, mhlanga2021artificial, di2020artificial, hajikhani2022mapping, holloway2018spatial,hall2023review}.

\section{Methods}
\label{sec:methods}

\subsection{Ground-truth data}
We use two types of surveys available through the Demographic and Health Surveys Program~\cite{dhs}: the Standard Demographic and Health Survey (DHS) and the Malaria Indicator Survey (MIS). Both are nationally-representative population-based surveys conducted at a household level. In particular, we focus on the housing characteristics questionnaire, which helps estimate the wealth of a household by considering the quality and quantity of available facilities or assets at home. In these surveys, household respondents are anonymized and assigned to a geo-located \textit{cluster} from which we can compute mean wealth values and their respective standard deviations. For simplicity, we refer to both surveys as DHS surveys.

\para{Cluster locations.}
\label{sec:clusters}
Clusters are typically census enumeration areas selected with probabilities proportional to the size within each stratum~\cite{dhscluster}, and often contain about 25-30 households. Among other attributes, clusters possess a type of settlement (urban/rural) and are geo-located with added noise %
to protect the exact location of respondents~\cite{dhsanon}. The displacement is random and varies according to the type of settlement: Up to $2$ {Km} for urban and up to $5$ {Km} for rural clusters. A further $1\%$ of places in the latter group are displaced by a maximum of $10$ {Km}. Note that most DHS clusters are located in rural areas, $66\%$ in SL and $76\%$ in UG, see~\Cref{tbl:locations}.

\para{International wealth index (IWI).}
The IWI score is a comparable asset-based index of household’s material well-being, or economic status, that can be used for all low and middle income countries~\cite{smits2015international}. 
The IWI score for each household is computed using the answers of 10 questions taken from the housing characteristics or living standards questionnaire. Then, asset weights are derived using principal component analysis (PCA)~\cite{hotelling1933analysis} and re-scaled to achieve IWI scores between 0 and 100.
Households with IWI=0 have no assets and possess the lowest quality housing. In contrast, households with IWI=100 represent the richest end of the spectrum.
The advantage of IWI scores over other wealth indices provided by the DHS, is that IWI scores are comparable across countries and years.

\subsection{Populated Places}
In order to obtain a high-resolution poverty map of a given country, we collect all its populated places and use them as target locations to infer their IWI scores with a selected model. %
Populated places are all cities, towns, neighborhoods, villages, hamlets and isolated dwellings that exist in OpenStreetMap (OSM)~\cite{osm,osmpplace}, an online crowd-sourced platform that provides annotations or metadata of the entire world. As in the ground-truth data, most populated places are located in rural areas, $96\%$ in SL and $99\%$ in UG, see~\Cref{tbl:locations}.

\subsection{Features}
\label{sec:features}
For each location, given by the centroid %
of each %
cluster and populated place, we extract 957~features from 7~data sources described below (see~\Cref{tbl:features} and~\cite{espin2022codefinal} for technical details). The default bounding-box width or radius used to query all features within a location is $1$ mile ($\approx$ $1.6$ Km). In some cases, we also query features within $2.0$, $5.0$, and $10.0$ Km to capture the original %
location of clusters, %
see~\Cref{sec:clusters}. 
All features were extracted in June 2022.

\para{Type of settlement.} Rural areas often host the poorest population of a country, and despite being characterized by their relative abundance of natural resources, they lack good quality of services and infrastructure~\cite{wiggins2001special}. 
Thereby, we distinguish each cluster according to the urban and rural divide by using a flag provided by the DHS.
In the case of populated places, we follow OSM's standards~\cite{osmpplace} and define urban places as cities, towns, and neighborhoods, and rural places as villages, hamlets, and isolated dwellings.

\para{Daylight satellite images.} We download the daylight satellite image of each place using the Google Maps Static API~\cite{staticapi}. These images are recorded in RGB bands with the resolution of 640x640 pixels corresponding to %
$\approx$ $2.5$ m per pixel. We normalize all pixels, 
and remove the logo and copyright label to prevent erroneous generalizations.\footnote{We crop the image by 620x620 pixels starting from the top-left corner.} 
The result image covers an area of $\approx$ 1.6x1.6 Km$^2$. 

\para{Population.} High-resolution population density maps are obtained through the ``Data for Good'' platform by Meta~\cite{fbpopulation}. In a nutshell, using state-of-the-art computer vision techniques, these maps estimate the number of people living within $30$ m grid tiles in nearly every country around the world~\cite{tiecke2017mapping}. Using these maps, we build \textit{9 population-based features}, see~\Cref{tbl:features}.
The ``gravitational'' features are motivated by the literature on population dynamics~\cite{khavinson2017gravitational}, and the selected $\beta$ exponents are typical values found in real-world datasets~\cite{barthelemy2011spatial}.
We use the ``closest tile'' information and ``total population'' within different radii for each location to overcome the different resolutions between our locations and the population tiles.

\para{Mobility.} Movements between tiles are also obtained through Meta's ``Data for Good'' platform~\cite{maas2019facebook}. These maps aggregate movement counts per day in three 8-hour windows %
to provide quick and localized responses after any natural disaster, and health crisis. %
The data provides the number of people moving between tiles at a given day and time-window as well as a \textit{baseline} reflecting the average movement between the same tiles, weekday, and time-window before the crisis. %
The resolution of these tiles is roughly 600x600 m$^2$ at the equator %
and they do not necessarily map the population tiles.
We use the baselines from the COVID-19 maps, %
and generate mobility networks where every tile is a node, and every movement %
is an edge in the graph. In SL (UG) we found 544 (2.5K) nodes, 2.9K (10.8K) edges, and 238K (1.2M) movements. 
First, we assign to each location %
the closest tile, and then derive %
\textit{27 mobility-based features} (see~\Cref{tbl:features}). 
Note that while all these features align with previous findings on the relationship between the wealth in Sub-Saharan African countries and mobility patterns %
(frequency, distances, modalities, and purpose)~\cite{bryceson2003livelihoods}, the ``movements between tiles'' dataset has not been used earlier to predict poverty maps.

\para{Demographics.} Using the Marketing API by Meta~\cite{fbmarketing}, we extract the number of monthly active users (MAU) on Facebook. Precisely, for each location %
we make $37$ independent queries to obtain the number of MAU that match certain demographics or characteristics and whose home location lies within a radius of $1.6$ Km. Motivated by previous work~\cite{fatehkia2020mapping,fatehkia2020relative,giurgola2021mapping}, the idea is to use these \textit{37 demographic-based features} as proxies of wealth, see~\Cref{tbl:features}. For instance, one hypothesis may claim that \textit{the more smartphone and tablet owners in the area, the wealthier the users living in there}.

\para{Infrastructure.} Similar to the daylight satellite images, the infrastructure features capture the development of a place. Their main difference is that while infrastructure features are collected via online crowd-sourcing, %
satellite images are remotely sensed. Further, infrastructure features are computationally easier to handle, however, they often provide outdated or no information about unpopular places. Here, we collect \textit{54 infrastructure-based features} from OSM~\cite{osm,osminfra,Mocnik_Open_Source_Data_2018}, e.g., number of buildings or ATMs within a 1.6x1.6 Km$^2$ bounding box around each location, and the distance to the closest road or a particular point of interest, see~\Cref{tbl:features}. The advantage of the ``closest distance'' features (without bounding boxes) is to help mitigating the missing data and unpopularity issues (e.g., the poorer the area, the longer the distance to the closest infrastructure).
Note that although OSM features can be queried with a specific date, this date does not necessarily reflect the time since the feature has been physically available; instead, it shows its \textit{last modified date} in the %
platform. Thus, we omit \textit{time} in the query.

\para{Connectivity.} Another important %
proxy of wealth is provided by the number of antennas within an area, and the distance to the closest tower. For each location, we construct \textit{9 connectivity-based features} extracted from OpenCelliD~\cite{opencellid}, see~\Cref{tbl:features}. As before, the \textit{distance} features help us mitigate missing values.

\begin{figure*}[t]
    \centering
    \includegraphics[width=0.88\textwidth]{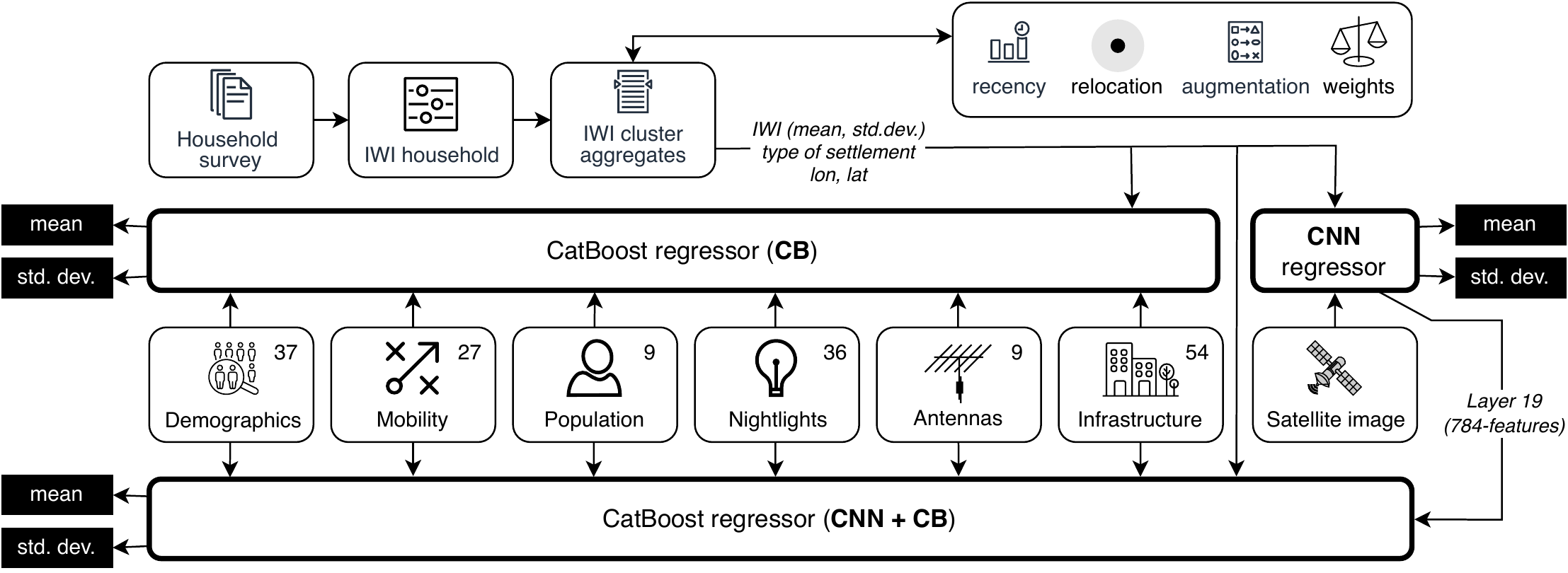}
    \caption{\textbf{Architecture.} We propose three models that learn to predict both the mean wealth of places and their respective standard deviation. The former represents the average wealth of households in a given area, and the latter indicates by how much the wealth of local households deviates from the mean. The Convolutional Neural Network model (CNN) is trained with daylight satellite images, and the CatBoost model (CB) is trained with metadata features such as demographics and mobility. The CNN+CB model is an extension of CB where in addition to the metadata features, it includes the third-to-last layer of the CNN as a low-dimension feature vector representation of the satellite images.
    Pre-processing of ground-truth data includes the computation of IWI scores per household, aggregation of IWI scores per cluster, and configuration of train and test sets to control for data recency, noisy locations, small training samples, and class imbalance.
    }
    \label{fig:model}
\end{figure*}

\para{Nightlight intensity.} The intensity of light at night has been shown to be a good predictor of wealth, especially in wealthy areas~\cite{pinkovskiy2016lights,jean2016combining,lee2020high,uzkent2021gen}. We leverage these correlations and construct \textit{36 nightlight-based features} for each location using the monthly average radiance composite images extracted from Google Earth Engine~\cite{eartheng,gee_viirs}. %
In particular, we measure 9 statistics within 4 different {buffers}, see~\Cref{tbl:features}.
We apply a mask to all images and retrieve all pixels whose average radiance 
is greater than or equal to a threshold, $\vn{rad}\geq 10$, and compute the respective fraction of pixels, area, and radiance that fulfill that criteria. Though this threshold is arbitrary, it serves as a proxy to exclude empty areas. Finally, we rank all pixels based on their radiance and compute the mean among the top-30\% and lowest-30\%.
Note that this is the only data source that supports querying data from the past. Thus, in the case of DHS clusters, we make sure that the year of the nighttime data matches the year of the survey, and for populated places we query the current year. Additionally, we standardize each metric per year. %

\subsection{Models}
\label{sec:models}
We analyze the distribution of wealth across households \textit{in each cluster}, and found that around 82\% of them are normally distributed.\footnote{Unlike the power-law wealth distribution of entire populations.}
Thus, we propose three machine learning models that learn to predict both the \textit{mean} and \textit{standard deviation} of IWI scores, see~\Cref{fig:model}.
By predicting these two values, we gain additional information on how much the wealth of households deviates from each other in each cluster. In addition, using the inferred distributions allows to build synthetic populations which are crucial inputs for development or epidemic modeling.
For implementation details, see~\cite{espin2022codefinal}.

\para{Image-based model (CNN).}
Our first model learns to predict IWI scores using daylight satellite images, see~\Cref{fig:model}.
Inspired by~\cite{naveed2019cnn}, this Convolutional Neural Network model contains 22~layers whose final-layer activation function is \textit{linear} and uses the mean-squared-error (MSE) as loss function.
Here, we tune $4$ hyper-parameters. %

\para{Metadata-based model (CB).}
Our second model is a CatBoost regressor model~\cite{prokhorenkova2018catboost} that learns to predict IWI scores from 173~metadata features, see~\Cref{fig:model,sec:features}. 
We additionally run this model for each data source separately to evaluate their individual performance. %
This model is tuned by $11$ hyper-parameters. %

\para{Combined model (CNN+CB).}
Our third model %
feeds the third-to-last layer of the CNN (layer \#19) into the CB model, as additional 784~features, see~\Cref{fig:model}.
Here, we verify whether all 957~(metadata and image-based)  
features %
produce the highest performance.

\subsection{Training}
\label{sec:training}
We partition the data into train (80\%) and test (20\%) sets, and stratify these partitions with respect to a 10-class discretized value of wealth (i.e., dividing IWI scores into 10 equal-width bins).
Note that these %
classes are used for sampling purposes only, 
thus, they do not intervene in training or inference (see~\Cref{sec:limitations} for a discussion on sampling alternatives). %
We further use the train set for a 4-fold cross-validation, %
and tune the hyper-parameters of each model via Random Search on 200 combinations. 
We then use the best combination of hyper-parameters to train the model on the entire train set. %
To control for random fluctuations, we repeat this procedure 3 times (i.e., in each run we shuffle the data and use different random seeds)
and report the mean performance. %
Note as well that %
we evaluate our models on different configurations of training data: (i) We use different years of ground-truth data to study data recency issues, (ii) compare different relocation strategies to evaluate performance on noisy locations, (iii) compare the performance of the CNN model with and without data augmentation, (iv) compare multiple sample weight strategies to mitigate wealth imbalance, and (v) compare the performance of our models across different feature sources.
Thus, for %
comparison, the split of the data 
sets is the same across relocations of the same recency and run.

\section{Results}

\subsection{Configuration trade-offs}

While designing our models, we identified several common ground-truth data issues that can undermine %
the wealth inference problem. Their systematic exploration provides our first set of contributions. %

\para{Data recency.} When using multiple data sources, a possible complication comes from the different times when the datasets are recorded. For instance, most features are derived from snapshots taken in 2022,
while the most recent available surveys date back to 2019 and 2018 for SL and UG, respectively. %
Additionally, the most recent surveys only represent a few clusters, which might not be sufficient to train generalizable machine learning models. %
To mitigate the trade-off between sample size and data recency, we first evaluate our models on different temporal configurations of training and testing data using the latest two surveys as follows. (i) O-O: trains and tests on 2016, the \textit{oldest} data. (ii) N-N: as before but on the \textit{newest} data, 2018 in UG and 2019 in SL. (iii) O-N: trains on the \textit{oldest} data, and tests on the \textit{newest}. (iv) ON: combines both years in the train and test sets. 
We apply stratified sampling in (i), (ii), and (iv), see~\Cref{sec:training}.
\Cref{fig:baselines} (a) and (e) show the results of recency for SL and UG, respectively. 
We found that the best performance for predicting the mean IWI in both countries is achieved by ON, while N-N and O-O are best for predicting the standard deviation in SL and UG, respectively, followed by ON.
Thus, we choose ON as a baseline for the upcoming experiments since overall provides the best performance in both countries with a larger sample.
For demonstration, %
we developed an interactive tool~\cite{yang2022vizfinal} to compare the inferred IWI scores at two different points in time.

\para{Relocation.}
Geo-located ground-truth data is anonymized by added noise to each cluster location to preserve the confidentiality of survey respondents~\cite{dhsanon}. While this is fundamental for ethical reasons, it induces uncertainty in the predictions. Previous work addresses this issue by either covering bigger areas around each cluster~\cite{Chie2022micro}, or re-arranging locations in an iterative way while also adjusting their wealth~\cite{lee2020high}. While these options are plausible, it is unclear by how much they affect the prediction. Here, we move the noisy location of a cluster to the location of the closest populated place (obtained from %
OSM) without altering its IWI score. %
In case multiple clusters are assigned to the same populated place, we prioritize the cluster with fewer other potential matches. We repeat this iteratively until all clusters are re-arranged. We test two alternatives of this algorithm: (i) rc: where only rural clusters are relocated to their closest rural populated place, and (ii) ruc: where both rural and urban clusters are relocated accordingly. 
\Cref{fig:baselines} (b) and (f) show the results of relocation. 
In UG it is best to keep the noisy locations, while in SL performance improves when all or only rural clusters are relocated, though the gain compared to no relocation is minor. Thus, we keep the noisy locations for the upcoming experiments.

\begin{figure}[ht!]
     \centering
\includegraphics[width=0.48\textwidth]{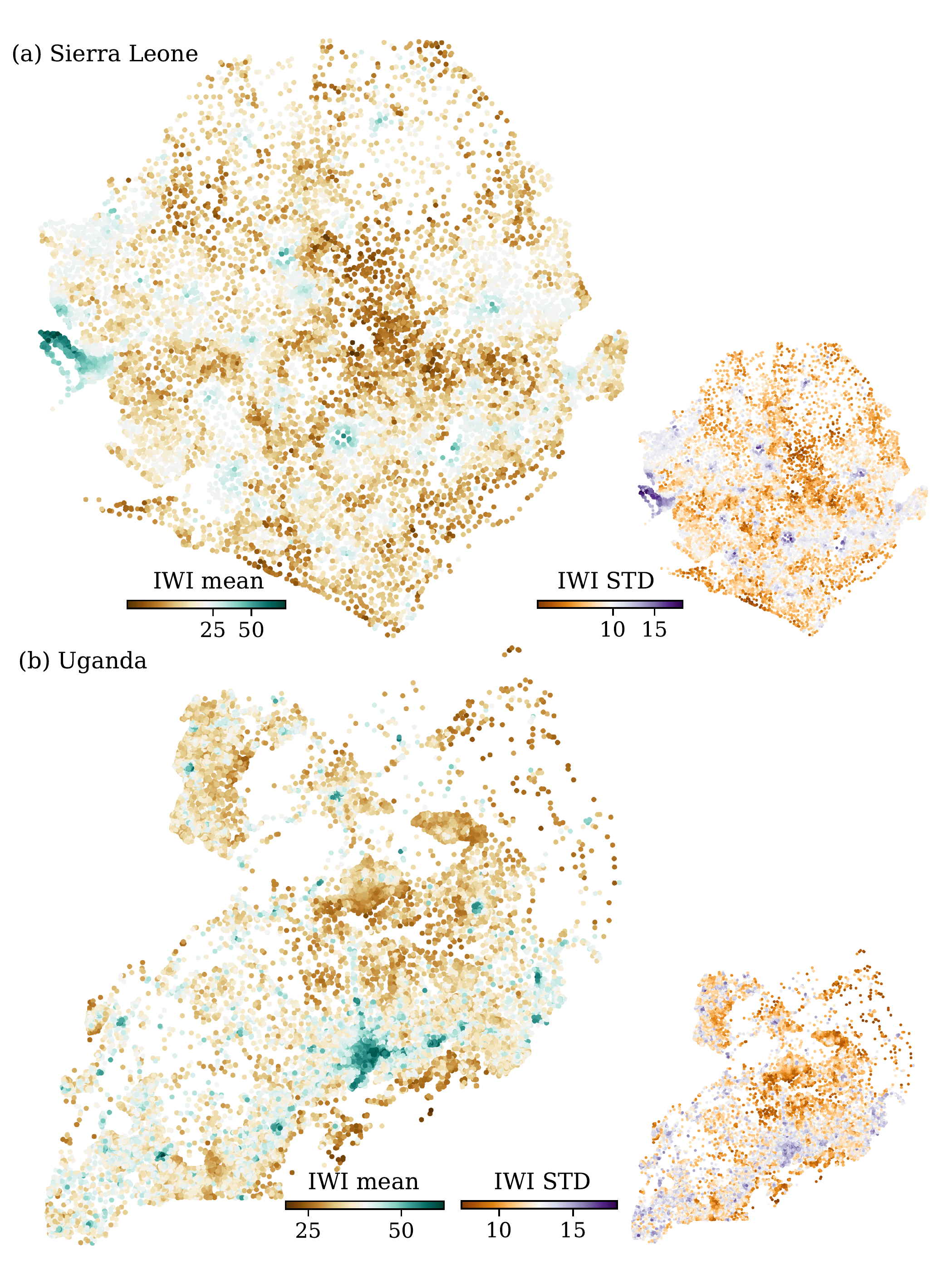} 
\vspace{-0.6cm}
    \caption{\textbf{High-resolution poverty maps.} Inferred mean (left) and standard deviation (right) of IWI scores %
    using the best models, CB$_w$ and CB, at the populated places of (a) SL and (b) UG, respectively.
    Maps on the left (right) are color-coded from brown-poor (orange-homogeneous) to green-rich (purple-heterogeneous) centered on their respective means. 
    The capitals of both countries are among the wealthiest and more heterogeneous areas. Also, SL appears poorer than UG, which aligns with the estimated poverty rates by the World Bank~\cite{WPR22}: 56.80\% in 2018 for SL, and 21.40\% in 2016 for UG. }
    \label{fig:poverty_maps}
    \vspace{-0.2cm}
\end{figure}

\para{Sample size.}
The performance of deep neural networks often improves with larger amounts of training data, and the $\approx$ $1000$ data points %
obtained by adding up the last two available surveys (see~\Cref{tbl:locations}), might still not be enough.
Thus, we apply offline augmentation during training to increase the training samples by 500\%. 
In particular, for each cluster in the image-based model (see~\Cref{sec:models}), we generate 5 new images as modifications of the original satellite image by applying different augmentation techniques.\footnote{brightness, noise, crop, erase, and rotation or flips.}
Unfortunately, our experiments suggest that augmentation adds little to nothing to the performance, see~\Cref{fig:baselines} (c) and (g). However, this can be improved by %
considering more sophisticated techniques~\cite{naveed2021survey, shorten2019survey}, which we leave for future work.

\begin{table*}[t]
\centering
\caption{
\textbf{Performance by model and feature-source.}
Each row shows the 
normalized root-mean-squared error (NRMSE$=\epsilon$) of predicting the mean $\mu$ and standard deviation $\sigma$ of IWI scores. 
The CatBoost model (CB) is trained with metadata features (i.e., Demographics, Mobility, Population, Nightlight intensity, Antennas, and Infrastructure), and the CNN model (CNN) with daylight satellite images.
The combined model (CNN+CB) uses all features. %
Model sub-indexes refer to ``weighted samples'' (CB$_w$) and ``offline augmentation'' (CNN$_a$), two techniques applied during training to correct for wealth imbalance and small samples, respectively.
The best performance across models is shown in \textit{blue}, and within each model type in bold.}
\label{tbl:model}
\begin{center}
\begin{tabular}{@{}llc@{\hskip7pt}c@{\hskip7pt}c@{\hskip7pt}c@{\hskip7pt}c@{\hskip7pt}c@{\hskip7pt}ccrrcrr@{}}
\toprule
 && \multicolumn{7}{c}{{Features}} & \phantom{a} & \multicolumn{2}{c}{{Sierra Leone}} & \phantom{a} & \multicolumn{2}{c}{{Uganda}} \\ \cmidrule{3-9} \cmidrule{11-12} \cmidrule{14-15} 
 & & De & Mo & Po & Nl & An & In & Im && {$\epsilon_{\mu}$} & {$\epsilon_{\sigma}$} && {$\epsilon_{\mu}$} & {$\epsilon_{\sigma}$} \\ \midrule
Metadata-all & CB & \checkmark & \checkmark & \checkmark & \checkmark & \checkmark & \checkmark & - && 0.46 & 0.81 && \best{0.46} & \best{0.83} \\
& CB$_w$ & \checkmark & \checkmark & \checkmark & \checkmark & \checkmark & \checkmark & - && \best{0.44} & \best{0.78} && 0.47 & 0.85 \\ \hline
Metadata-single & CB$_w$ & \checkmark & - & - & - & - & - & - && 0.73 & 0.94 && 0.77 & 0.97 \\
&  & - & \checkmark & - & - & - & - & - && 0.47 & 0.81 && 0.55 & 0.92 \\
&  & - & - & \checkmark & - & - & - & - && \textbf{0.45} & \textbf{0.79} && 0.61 & 0.90 \\
&  & - & - & - & \checkmark & - & - & - && 0.52 & 0.83 && \textbf{0.50} & \textbf{0.86} \\
&  & - & - & - & - & \checkmark & - & - && 0.52 & 0.87 && 0.56 & 0.89 \\
&  & - & - & - & - & - & \checkmark & - && 0.51 & 0.85 && 0.61 & 0.89 \\ \hline
Image-only & CNN & - & - & - & - & - & - & \checkmark && \textbf{0.57} & \textbf{0.90} && 0.60 & 0.94 \\
& CNN$_a$ & - & - & - & - & - & - & \checkmark && 0.59 & 0.94 && \textbf{0.59} & \textbf{0.91} \\ \hline
Combined & CNN+CB & \checkmark & \checkmark & \checkmark & \checkmark & \checkmark & \checkmark & \checkmark && 0.45 & 0.82 && 0.51 & 0.86 \\
& CNN+CB$_w$ & \checkmark & \checkmark & \checkmark & \checkmark & \checkmark & \checkmark & \checkmark && \textbf{0.45} & {0.80} && 0.52 & 0.86 \\
& CNN$_a$+CB & \checkmark & \checkmark & \checkmark & \checkmark & \checkmark & \checkmark & \checkmark && 0.46 & 0.82 && \textbf{0.47} & \textbf{0.83} \\
& CNN$_a$+CB$_w$ & \checkmark & \checkmark & \checkmark & \checkmark & \checkmark & \checkmark & \checkmark && 0.45 & \textbf{0.79} && 0.47 & 0.85 \\
\bottomrule
\end{tabular}
\end{center}
\end{table*}

\para{Class imbalance.}
A further complication comes from imbalanced ground-truth data. 
In terms of types of settlements, both clusters and populated places are mostly located in rural areas, see~\Cref{tbl:locations}.
Wealth is also unequally distributed across clusters (i.e., $Gini\approx0.31$ for both countries\footnote{The Gini coefficient is a measure of income inequality in society~\cite{gini}.}), making the poor areas the majority.
Recall that IWI scores are continuous real values that range from 0 (poorest) to 100 (richest).
In order to account for socioeconomic class imbalances we first discretize these scores into 10 equal-width bins. %
Then, we apply a class-balanced loss based technique~\cite{cui2019class} to generate a weight for each data point in the training sample and pass these weights to the classifier.\footnote{
We control the imbalance issues on wealth and type of settlement separately and in combination using 4 techniques: heuristics, Inverse Number of Samples (INS), Effective Number of Samples (ENS), and $\vn{compute\_sample\_weight}$ from $\vn{sklearn}$ in python~\cite{sklearn22csw}. 
Due to space limitations, out of the 20 configurations, we report results using ENS with $\beta=0.9$, the technique with best performance, see~\cite{espin2022codefinal} for more details.}
We see in~\Cref{fig:baselines} (d) and (h) that adding sample weights only benefits the metadata model in SL.

\subsection{Performance}
Our second contribution and main outcome of our methodology is the inferred high-resolution poverty maps of SL and UG, see~\Cref{fig:poverty_maps}. These maps are explained in more detail, for all investigated models, in our online interactive tool~\cite{yang2022vizfinal}.

Our third and last set of contributions addresses our research questions.
Here, we disentangle the performance of our models by reporting the normalized root-mean-squared errors ($NRMSE=\frac{RMSE}{\sigma_{y\_\text{true}}}$) for the inference of both predicted mean and predicted standard deviation (STD). Results are based on the combination of the last two available surveys %
using noisy locations. %
For comparison, we report model performance with and without augmentation (CNN$_a$ vs. CNN), and with and without sample weights (CB$_w$ vs. CB).

\para{RQ1: Model selection and feature-source importance.}
As shown in~\Cref{tbl:model}, we find that the \textit{metadata-all} models, which use all non-image features, outperform the other models in terms of predicted mean ($\epsilon_\mu$) and STD ($\epsilon_\sigma$) of IWI scores. In SL, the CB$_w$ model, that balances the skewness of wealth in the data, provides the best performance, while the unbalanced CB model appears the best in UG (shown in \textit{blue}). Interestingly, including all features in the combined model improves the performance only for UG as compared to the performance of individual feature-sources.

From the \textit{metadata-single} results shown in~\Cref{tbl:model}, we find that while most of the features can provide relatively good predictive performance, the population (Po) and nightlight intensity (Nl) features alone are good predictors of wealth in SL and UG, respectively. This aligns with previous studies that show that population density~\cite{bettencourt2013origins} and luminosity~\cite{jean2016combining} are strongly correlated with economic growth.
Surprisingly, the \textit{image-only} models achieve the worst performance, however, they can be improved by \textit{combining} them with the metadata features.

\begin{table*}[t]
\caption{
\textbf{Performance by intersectionality.} 
For each country we divide the true mean IWI scores into %
quintiles and measure
the root-mean-squared error (RMSE) of predicting the mean wealth
($\epsilon_\mu$) 
at the intersection between countries, types of settlements, quintiles, and models.
Bold values refer to the best performance in each quintile and type of settlement.
}
\label{tbl:subpop}
\begin{center}
\begin{tabular}{@{}llrrrrrcrrrrr@{}}
\toprule
 & & \multicolumn{5}{c}{{Sierra Leone}} & \phantom{a} & \multicolumn{5}{c}{{Uganda}} \\ \cmidrule{3-7} \cmidrule{9-13} 
& \multicolumn{1}{c}{} & \multicolumn{1}{r}{{Q1}} & \multicolumn{1}{r}{{Q2}} & \multicolumn{1}{r}{{Q3}} & \multicolumn{1}{r}{{Q4}} & \multicolumn{1}{r}{{Q5}} && \multicolumn{1}{r}{{Q1}} & \multicolumn{1}{r}{{Q2}} & \multicolumn{1}{r}{{Q3}} & \multicolumn{1}{r}{{Q4}} & \multicolumn{1}{r}{{Q5}} \\ \midrule
Rural & CB              &  \textbf{5.48} & \textbf{3.18} &   5.77 &  10.46 &  23.20 &&   7.01 &   \textbf{4.06} &   3.63 &   8.04 &  \textbf{12.54} \\
& CB$_w$          &  7.64 &   4.83 &   5.18 &  10.18 &  22.47 &&   8.83 &   5.08 &   5.06 &   8.47 &  15.48 \\
& CNN             &  7.72 &   4.48 &   \textbf{4.67} &  10.59 &  22.56 &&   9.84 &   5.51 &   4.87 &   7.55 &  16.47 \\
& CNN$_a$         &  8.96 &   5.07 &   4.80 &  10.30 &  23.41 &&   9.66 &   5.11 &   4.61 &   \textbf{7.44} &  16.72 \\
& CNN+CB          &  6.24 &   3.49 &   4.78 &   9.81 &  \textbf{20.18} &&   7.20 &   4.49 &   4.26 &   7.67 &  15.07 \\
& CNN+CB$_w$      &  6.23 &   3.66 &   4.75 &   \textbf{9.74} &  21.36 &&   7.67 &   4.51 &   4.22 &   7.59 &  15.07 \\
& CNN$_a$+CB      &  6.74 &   3.38 &   5.03 &  10.55 &  22.64 &&   \textbf{6.67} &   4.26 &   \textbf{3.35} &   7.54 &  13.78 \\
& CNN$_a$+CB$_w$  &  6.55 &   3.45 &   5.02 &  10.54 &  22.11 &&   6.87 &   4.15 &   3.59 &   7.59 &  13.84 \\ \hline
Urban & CB              &     - &  18.07 &   8.13 &   6.18 &   9.69 &&   \textbf{6.42} &   6.96 &  10.08 &   6.77 &   \textbf{9.33} \\
& CB$_w$          &     - &  10.30 &   8.16 &   8.67 &  11.34 &&  11.70 &   7.38 &  12.19 &   9.68 &  12.83 \\
& CNN             &     - &   \textbf{4.54} &   2.83 &  11.03 &  12.84 &&   9.77 &   5.41 &  11.95 &   7.77 &  13.38 \\
& CNN$_a$         &     - &   6.77 &   \textbf{2.21} &  10.96 &  13.23 &&  10.11 &   \textbf{5.10} &  13.94 &   8.34 &  13.18 \\
& CNN+CB          &     - &  15.48 &   4.73 &   6.70 &   9.62 &&   9.02 &   5.99 &  10.63 &   7.71 &  10.97 \\
& CNN+CB$_w$      &     - &  14.15 &   6.53 &   6.77 &   9.33 &&   8.66 &   7.13 &  10.96 &   8.13 &  11.08 \\
& CNN$_a$+CB      &     - &  16.17 &   6.84 &   5.82 &   9.48 &&  10.45 &   5.77 &  10.03 &   \textbf{6.67} &   9.67 \\
& CNN$_a$+CB$_w$  &     - &  15.99 &   8.06 &   \textbf{5.75} &   \textbf{9.31} &&   8.86 &   6.54 &   \textbf{9.86} &   6.72 &   9.70 \\
\bottomrule
\end{tabular}
\end{center}
\end{table*}

\para{RQ2: Intersectionality.}
The overall performance shown above is not informative about the intersectionality of the models, e.g., 
the prediction of wealth in poor-rural areas might not be as accurate as in rich-urban areas.
Therefore, we assess the performance of our models at the intersection between types of settlements and socioeconomic classes; both derived from the DHS data. The former is given, and the latter we infer by transforming the true mean IWI scores into quintiles (i.e., 5 equally-populated bins).
Results based on the predicted mean wealth ($\epsilon_\mu$) are shown in~\Cref{tbl:subpop}. We see that the CB models, which appeared to be the best in the overall performance (in~\Cref{tbl:model}), were found to be good only in certain regions. 
In SL, the lowest quintiles (Q1, Q2) are best captured by the CB model in rural areas, while the middle (Q2, Q3) and rich quintiles (Q4, Q5), regardless of their type of settlement, are best captured by the CNNs and the combined models, respectively.
In UG, on the other hand, the richest quintile (Q5) is best captured by the CB model, while the middle quintiles (Q2-Q4) are best explained by the CNNs or combined models everywhere. Moreover, the poorest quintile (Q1) is best captured by the combined and CB models in rural and urban areas, respectively.
These results demonstrate that there is no model that fits very well all types of settlements and wealth quintiles at the same time; thus, it is important to systematically explore them across different sub-populations.

\begin{figure}[b] %
    \centering
    \includegraphics[width=0.48\textwidth]{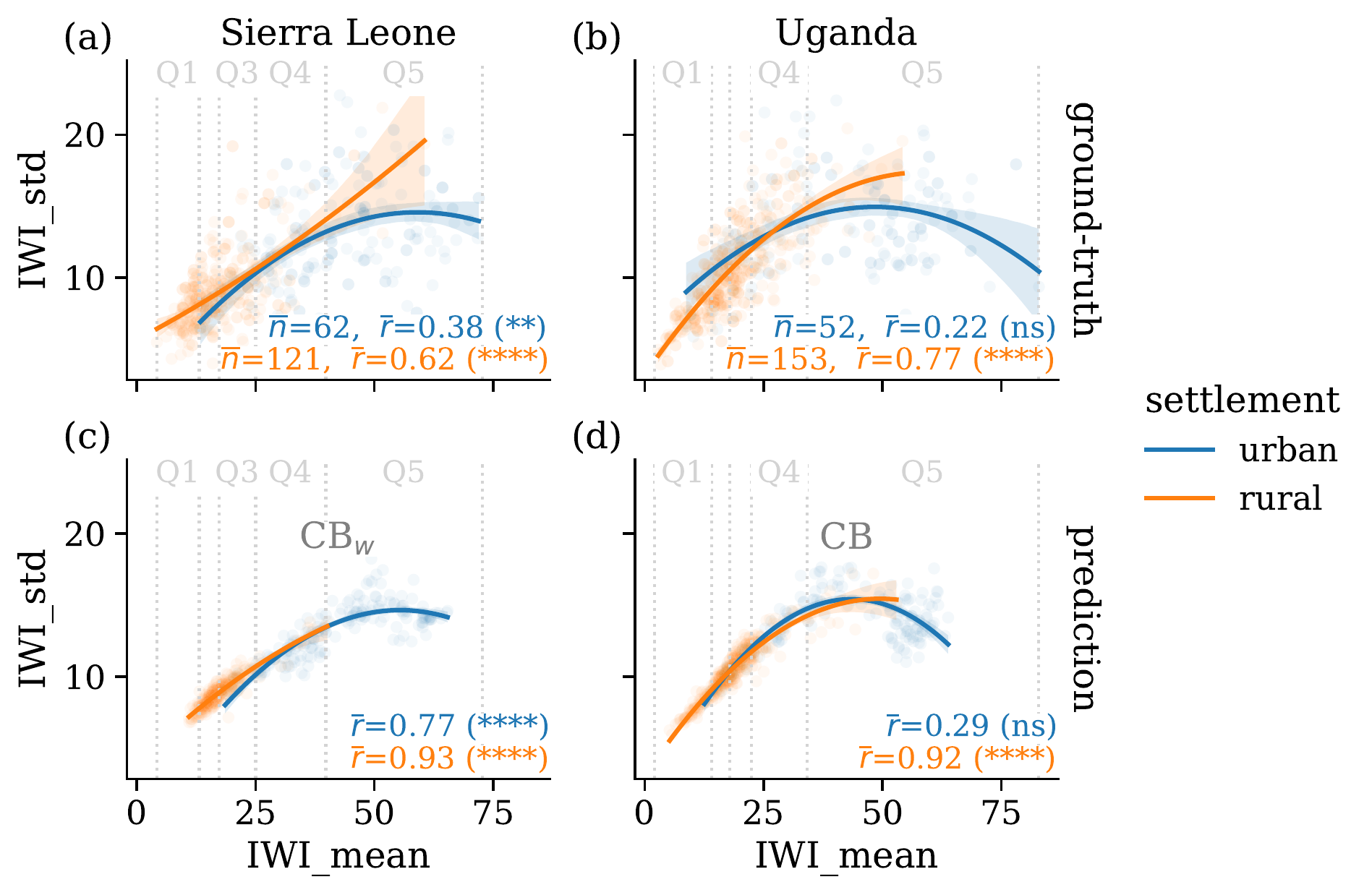}
    \caption{\textbf{Wealth variability.} 
    Relationship between the mean (x-axis) and standard deviation (y-axis) of wealth in SL and UG (columns) for urban (blue) and rural (orange) clusters on the test sets. %
    True values are shown at the top, (a) and (b), and predicted values using the best model in each country are at the bottom, (c) and (d).
    Curves are $2^{nd}$ order polynomial fits, and labels show the average \# of places $\bar{n}$, and the mean Pearson correlation $\bar{r}$ between the two predicted variables. %
    }
    \label{fig:variability}
\end{figure}

\para{RQ3: Wealth variability.}
When predicting aggregated economic indicators, mean values are often not enough to understand the distribution of wealth across households. 
Thus, by predicting the STD, and understanding its relation with the mean, %
we gain additional insights about the socioeconomic diversity of individuals living in the same populated places. %
From the results shown in~\Cref{fig:variability}, we find that the best inference model in each country predicts well---with slight overestimation---the minimum mean wealth %
as compared to the true values (orange curves in Q1);
crucial for policy making when targeting places in need. However, the inference somewhat underestimates the mean wealth at the richest end of the spectrum (Q5) as the models miss to capture the wealth of outlier rich locations; especially in urban areas (blue curves).

Meanwhile, we also find a positive and significant correlation between the mean and STD of wealth for both true and predicted values; especially in rural areas (orange curves). This trend indicates that larger socioeconomic variation characterizes wealthier places. However, in urban areas, this correlation tends to follow a non-monotonic pattern. While the STD appears to be the smallest at the poorest places, after reaching a maximum for middle class locations, it gets smaller for the richest areas. This means that while middle class people tend to live in places with higher socioeconomic variations, the poorest and the richest live in more homogeneous communities, signaling their segregation from the rest of the population.
In \Cref{fig:pplace_variability}, we further confirm the positive correlation between population size and inferred wealth in populated places; aligned with our feature importance analysis. %
Moreover we find a strong distinction between rural and urban areas, the latter appearing more populated and wealthier, yet depicting the already mentioned segregated mixing patterns, especially in SL.

\begin{figure}[t]
    \centering
    \includegraphics[width=0.48\textwidth]{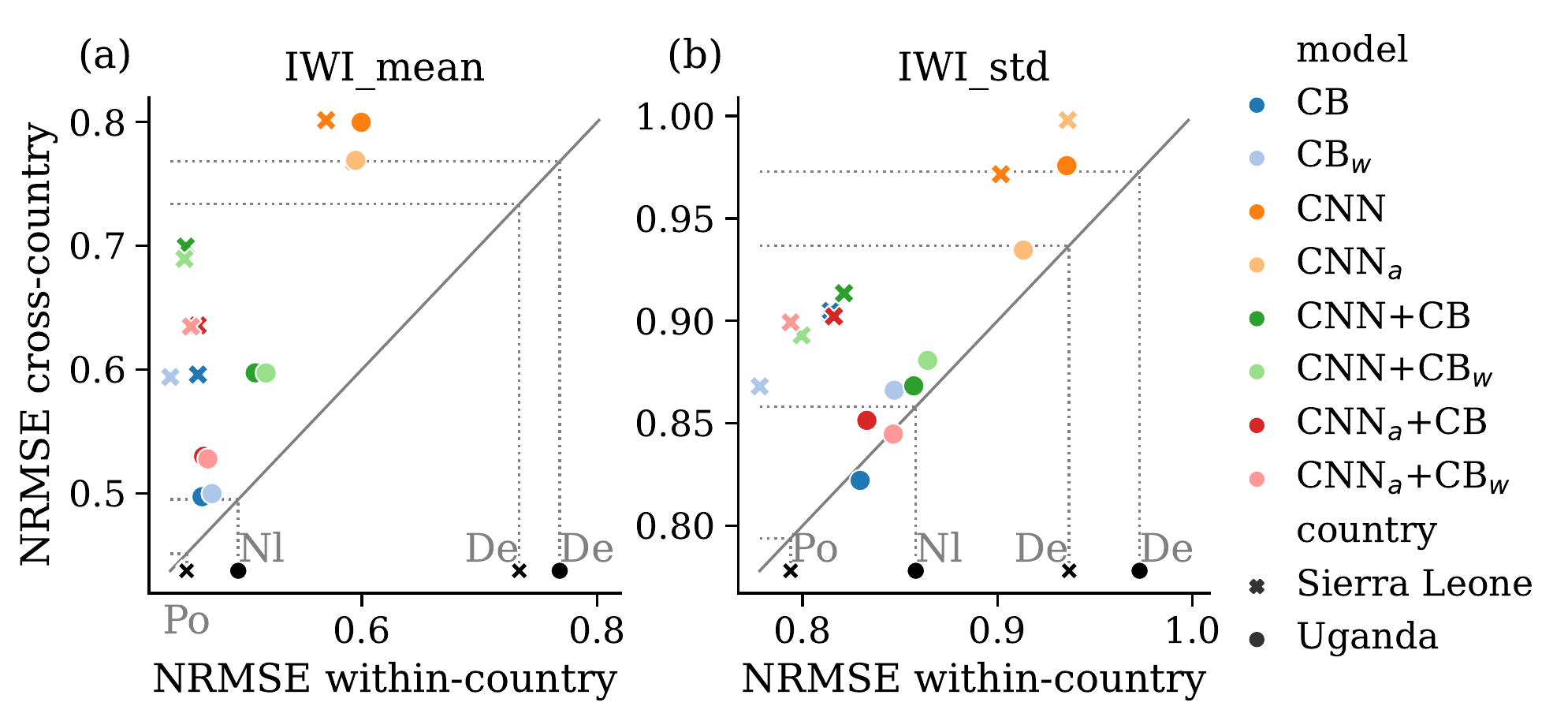}
    \caption{\textbf{Cross-country model performance.} 
    We test model transferability by using the models trained in one country to predict the wealth in the other without further training. 
    For each response variable %
    (columns a and b), model (color), and country (markers), we show model performance in terms of normalized root-mean-squared errors (NRMSE) for both within country (x-axis) and cross-country (y-axis) validations for SL ($\times$) and UG ($\boldsymbol{\cdot}$). %
    Dotted lines with labels indicate the within-country metadata-single performance for population (Po), nightlight intensity (Nl), and demographics (De).
    }
    \label{fig:cross}
\end{figure}

\para{RQ4: Cross-country model transferability.}
We test the transferability of our models by training them in one country to predict the wealth in the other without further training.
\Cref{fig:cross} shows the performance of the models in terms of normalized root-mean-squared errors (NRMSE) for both within country (x-axis) and cross-country (y-axis) validations. 
We found three main patterns: 
(i) Model transferability across countries achieves higher errors compared to the within-country counterpart, i.e., most data points are above the diagonal. Although this was expected, CB models provide surprisingly good performance in transferability, landing close to the diagonal line in many cases. 
This can be due to the several metadata sources these models rely on since they may compensate data quality differences across layers.
An interesting exception is for SL, where the STD is somewhat better predicted from a CB model trained on UG than the corresponding model trained on SL, see the dark blue dot below the diagonal in \Cref{fig:cross} (b).
(ii) Model transferability is asymmetric, i.e., training on SL to predict UG leads to smaller errors than vice versa. This may be due to the wider range of mean wealth in UG compared to SL, %
i.e., 99\% of clusters in UG fall within the distribution of mean wealth in SL, see~\Cref{tbl:locations}. 
Thus, models trained in UG may infer values that fall off the wealth range characterizing SL, resulting in larger inference errors.
(iii) There is a trade-off between model transferability and feature availability when only one feature source is present. Naturally, model transferability pays-off if no feature is available. %
However, if only the best metadata-single features are available, i.e., nightlight intensity (Nl) in UG and population (Po) in SL (leftmost dotted lines in \Cref{fig:cross}), then model transfer should be avoided since the transferred model's performance is worse than using the single available local feature source; except for CNN$_a$+CB$_*$ and CB when inferring STD in UG. On the other hand, if only the worst metadata-single features are available, i.e., demographics (De) (rightmost dotted lines in \Cref{fig:cross}), then model transferability yields an advantage for almost all models. %

\section{Limitations and future work}
\label{sec:limitations}
The main limitations of our methods are rooted in multiple aspects. 
First, the recency, quality, coverage, and biases of the different data sources that we use may hinder the performance of the inference task. We discussed in detail the effects of these issues and showed how to address them. Meanwhile, the multiple layers of collected features %
seem to account for the effects of low data quality in certain layers, making our models successful in predicting wealth even in data-sparse regions.
Second, even though our models achieve high performance, comparable with previous work~\cite{lee2020high,Chie2022micro}, we cannot assume the poverty maps to be the possibly most precise representation of actual wealth. As we argued, data recency and sample size are important aspects to consider during training. %
Moreover, the timestamps of most features do not match the timestamps of the ground-truth data; the former cannot be queried retrospectively in time. 
Thus, more recent ground-truth data is required to infer up-to-date poverty maps.
Third, the current implementation of our methodology most reliably works for low- and middle-income countries since it relies on DHS survey data. Nevertheless, our code and results are openly available~\cite{espin2022codefinal}, thus, new types of ground-truth data and features can be easily plugged into our framework, e.g., housing prices, mobile airtime payments, or street view imagery. %
Similarly, additional methods can be incorporated into our pipeline to extend the explanations of our results. For example, activation maps~\cite{abitbol2020interpretable} can be used to understand the under-performance of the image-only models in poor areas, and geographical stratification~\cite{Chie2022micro} could guarantee a better spatial coverage during training in addition to the stratification of wealth (cf.~\Cref{sec:training}).

Note that, while this and similar methodologies have been implemented for a social good, they may also be misused, e.g., for commercial purposes which may increase the inequality gap between the rich and the poor. One way to mitigate this issue is by implementing better synergies between data providers and analysts to keep track of whom is using the data and for what purposes.

In the future, we aim to tackle these limitations and explore the spatial and temporal transferability of single- and multi-source models with missing data layers at the target population. Further, we will consider combining the best of each model to produce more stable and consistent results across different sub-populations.

\section{Conclusion}
In this work, we showed how to produce high-resolution poverty maps to interpret wealth distributions using multimodal data.
We proposed machine learning models that learn to predict the mean and standard deviation of wealth in places from low- and middle income countries, and showcased their capabilities in Sierra Leone and Uganda. These models rely on a multimodal architecture composed by sensor and online crowd-sourced data. Moreover, we systematically studied four major issues namely data recency, noisy locations, small training samples, and class imbalance, and conducted an ablation study to evaluate their performance. 

Our findings show that:
(1) The combined data features outperform satellite images in solving the inference task. Further, population and nightlight intensity features are the strongest predictors of wealth in SL and UG, respectively, %
followed by the mobility features in both countries.
(2) There is no single model or feature source that can predict wealth very accurately at the intersection between all socioeconomic classes and types of settlements. However, in general, the poorest areas are best inferred by the CB models using metadata-only features, while wealthier areas are best inferred by the CNN or combined models using daylight satellite images.
(3) Our models capture well the mean and standard deviation of wealth, their correlation, and non-monotonous dependencies. This way they provide information about the socioeconomic heterogeneity at given places and reflect segregation between different classes.  %
(4) There is an asymmetric trade-off between geographical transferability and data availability where countries with sparse information %
may benefit from reusing models trained in other countries.

To conclude, our results shed light on the importance of producing accountable models to guarantee accurate predictions across all types of settlements and socioeconomic groups.
Finally, beyond our scientific results, we made our code and results openly available~\cite{espin2022codefinal}, and built an online interactive visualization tool~\cite{yang2022vizfinal} to help identifying places in need for the development of fair policy and intervention designs.

\begin{acks}
We thank Liuhuaying Yang for developing the visualization tool, and Gilles Vandewiele and Philipp Singer for their feedback.
We are also grateful for the support of the Data for Good Team at Meta, the SoBigData++ %
project (H2020-871042), Dataredux ANR project (ANR-19-CE46-0008), and the CHIST-ERA project SAI: FWF I 5205-N. 
This material is based upon work supported by the Google Cloud Research Credits program with the award GCP19980904.
The presented computational results have been achieved in part using the Vienna Scientific Cluster (VSC).
\end{acks}

\balance
\bibliographystyle{ACM-Reference-Format}

\onecolumn
\appendix

\setcounter{table}{0}
\setcounter{figure}{0}
\renewcommand{\thetable}{A\arabic{table}}
\renewcommand{\thefigure}{B\arabic{figure}}

\section{Appendix tables}

\begin{table}[h]
\centering
\caption{\textbf{Locations}. Household and cluster data are gathered via the Demographics and Health Survey program (DHS), and populated places from OpenStreetMap (OSM). Rural and urban labels of clusters are directly assigned by the DHS, and for populated places we infer them from their type: \textit{urban} if they are cities, towns or neighborhoods, and \textit{rural} if they are villages, hamlets or isolated dwellings. Statistical significance (****), {p-value} $\leq0.0001$.}
\label{tbl:locations}
\begin{tabular}{@{}lrr@{}}
\toprule
\textbf{Country} & \textbf{Sierra Leone} & \textbf{Uganda} \\ \midrule
Years                    &    2016, 2019 &    2016, 2018 \\
DHS households               &        19975 &        27798 \\
\textbf{DHS clusters}                 &          \textbf{893} &         \textbf{1001} \\
- oldest, newest  & 336, 557 & 685, 316 \\
- urban                    &    308 (34\%) &    242 (24\%) \\
- rural                    &    585 (66\%) &    759 (76\%) \\
- IWI mean ($\mu$)\\
\hspace{2mm}min, max, mean, std.dev.             &          4.2, 72.7, 26.2, 15.6  &          2.0, 82.9, 24.8, 14.9  \\
- IWI std.dev. ($\sigma$)\\
\hspace{2mm}min, max, mean, std.dev.              &          3.8, 22.8, 10.4, 3.5 &          3.9, 22.4, 11.4, 3.4 \\
- Pearson corr. ($\mu$ and $\sigma$)                 &  0.71 (****) &  0.64 (****) \\
- Gini IWI mean                     &         32 &         31 \\
WorldBank Gini index            &  35.7 (2018) &  42.7 (2019) \\
\textbf{OSM populated places}         &         \textbf{9881} &        \textbf{27791} \\
- urban            &     366 (4\%) &     348 (1\%) \\
- rural            &   9515 (96\%) &  27443 (99\%) \\
\textbf{Relocated DHS cluster}               &    \textbf{705} (79\%) &    \textbf{733} (73\%) \\
- urban         &    120 (39\%) &     71 (29\%) \\
- rural         &   585 (100\%) &    662 (87\%) \\
\bottomrule
\end{tabular}
\end{table}

\begin{table}[h]
\centering
\caption{\textbf{Features.} Population, mobility and demographics are datasets provided by Meta. Infrastructure features are extracted from OpenStreetMap. Connectivity features refer to the antennas provided by OpenCelliD. Nightlight intensity scores are provided by Google as well as the daylight satellite images. For more technical details see~\cite{espin2022codefinal}.}
\label{tbl:features}
\begin{tabular}{@{}llrl@{}}
\toprule
\textbf{Name} &
\textbf{Source} & \textbf{\#} & \textbf{Features} \\ \midrule
Population $^\dagger$ & Meta & 9 &  \begin{tabular}[t]{@{}l@{}}$\vn{distance\_to\_closest\_tile}$, $\vn{population\_in\_closest\_tile}$, $\vn{total\_population\_within(radius)}$, \\and $\vn{gravitational\_closest\_tile^\beta = population/distance^\beta}$, where $\beta\in\{1.0,1.5,2.0\}$ and \\$\vn{radius\in\{1.6, 2.0, 5.0, 10.0\}}$ Km.\end{tabular}  \\
Mobility $\ddagger$ & Meta & 27 &  \begin{tabular}[t]{@{}l@{}}$\vn{distance\_to\_closest\_tile}$, $\vn{average\_distance\_in}$, $\vn{average\_distance\_out}$, $\vn{people\_flow\_in}^\beta$, \\$\vn{people\_flow\_out}^\beta$, $\vn{in\_degree}^\beta$, $\vn{out\_degree}^\beta$, $\vn{pagerank}^\beta$, and $\vn{weighted\_pagerank}^\beta$, \\where $\beta\in\{\vn{None},1.0,1.5,2.0\}$. When $\beta=\vn{None}$, we assume raw values \\(without the $\beta$ exponent), otherwise we adapt the gravitational formula~\cite{khavinson2017gravitational} as before: \\$\vn{metric/distance}^\beta$, where  
the $\vn{distance}$ between the location 
and the closest tile \\is measured in meters using KD-tree nearest neighbor search~\cite{ram2019revisiting}.\end{tabular}\\
Demographics $\mathsection$ & Meta & 37 & \begin{tabular}[t]{@{}l@{}}Behavior (5 mobility, 1 business, 4 network, 6 technology), Life events (2), Industry \\(4 employment), User device (4 assets), User-OS (5 assets), Education (4), Interests (2)\end{tabular}\\
Infrastructure $\dagger$ & OpenStreetMap & 54 & Roads (4), Buildings (2), POIs (24 counts, 24 distances)\\
Connectivity $\dagger$ & OpenCelliD & 9 &  \begin{tabular}[t]{@{}l@{}}$\vn{distance\_to\_closest\_cell}$, $\vn{number\_of\_cells}\in radius$, $\vn{number\_of\_towers}\in radius$,\\where $\vn{radius\in\{1.6, 2.0, 5.0, 10.0\}}$ Km.\end{tabular}\\
Nightlight intensity $\ddagger$ & Google & 36 & \begin{tabular}[t]{@{}l@{}}$\vn{min}$, $\vn{max}$, $\vn{mean}$, $\vn{median}$, $\vn{frac\_pixels}$, $\vn{frac\_area}$,  $\vn{frac\_sum\_rad}$, $\vn{t30\_mean}$, and $\vn{l30\_mean}$ \\for $\vn{radius} \in \{1.6, 2.0, 5.0, 10.0\}$ Km.\end{tabular} \\ 
Daylight satellite images $^\mathsection$ & Google & - & A 640x640 image for each place at zoom level 16 (e.g., buildings and streets) \\\bottomrule
\end{tabular}
\end{table}
\vspace{-3mm}
\footnotesize{$^\dagger$ Open access  $^\ddagger$ Upon request $^\mathsection$ API costs can be covered by free-tiers, sand-box accounts, or Google Cloud research credits for Google products}\\

\clearpage
\onecolumn
\section{Appendix Figures}
\begin{figure}[h]
    \centering
    \includegraphics[width=1.0\textwidth]{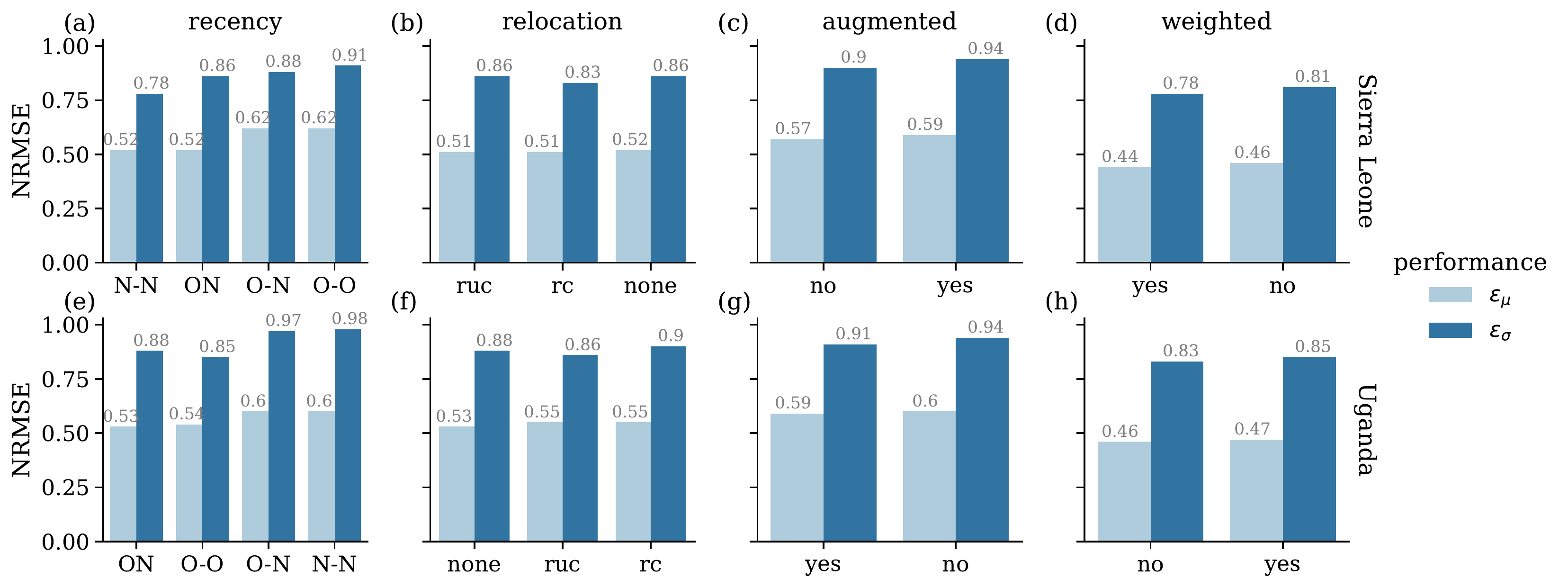}
    \caption{\textbf{Ablation results.} Results on (a,e) recency, (b,f) relocation, (c,g) augmentation, and (d,h) weighting are shown for SL and UG, respectively. Each bar shows the performance of each model as the normalized root-mean-squared error (NRMSE) for the prediction of the mean $\mu$ (light blue) and standard deviation $\sigma$ (dark blue) 
    of the IWI scores. 
    This normalization allows comparing the two outcomes, 
    and we see that $\mu$ is easier to infer than $\sigma$.
    Recency and relocation values are averages across the CB and CNN models (i.e., neither weighted nor augmented samples). 
    Additionally, experiments across different recency configurations use no relocation (relocation=none).
    The rest of experiments use the combined recency (recency=ON) since in both countries, combining the last two available surveys as ground-truth yields the smallest $\epsilon_\mu$ %
    across all recency configurations.
    In terms of relocation, the noisy locations are best for UG (relocation=none), and while relocation is better in SL (relocation$\ne$none), %
    we use the noisy locations in further experiments since the difference in performance is minimal.
    Augmentation only benefits UG, and weighted samples benefit only SL.
    Note that the combined models (CNN$_*$+CB$_*$) are excluded from this analysis.
    }
    \label{fig:baselines}
\end{figure}

\begin{figure}[h]
    \centering
    \includegraphics[width=0.65\textwidth]{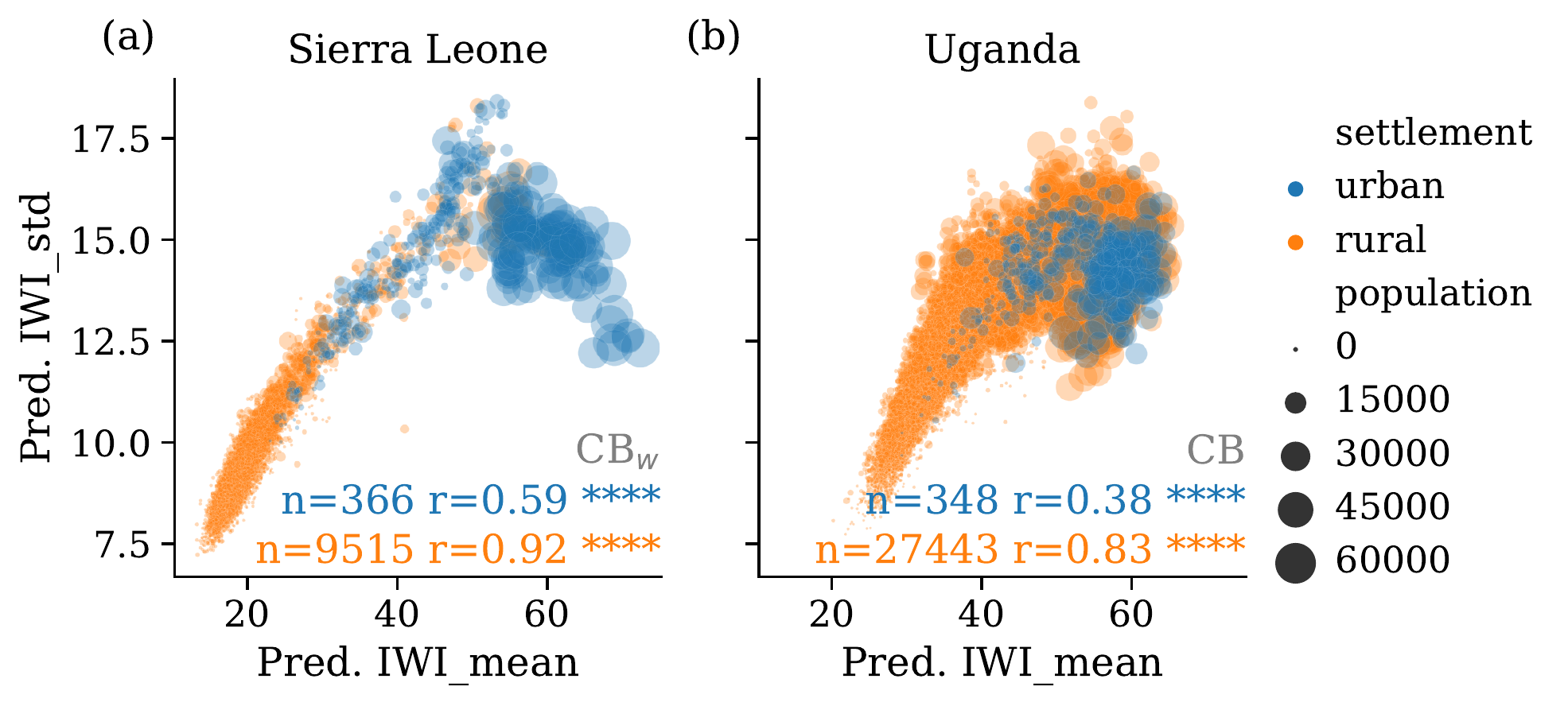}
    \caption{\textbf{Mean wealth, variability and population size:} These are the predicted values of wealth on the populated places of (a) SL and (b) UG using their respective best models CB$_w$ and CB. The x-axis shows the predicted mean IWI scores while the y-axis shows the predicted standard deviations. We see that urban places (blue) are often wealthier than the rural ones (similar as in the ground-truth data, see~\Cref{fig:variability}), and wealthier places tend to be more populated than the poor. Here, population refers to the number of total people within the populated place in a diameter of $d\approx1.6~\vn{Km}$ from its centroid. Annotations refer to the number of populated places $n$, and the Pearson correlation $r$ between the two predicted variables.}
    \label{fig:pplace_variability}
\end{figure}\end{document}